\documentclass{article}
\usepackage{spconf,amsmath,graphicx}

\usepackage{enumitem}
\setlist{nosep, leftmargin=14pt}

\usepackage{mwe} 
\usepackage{amsfonts} 
\usepackage{color}
\usepackage{booktabs}
\usepackage{comment}
\usepackage{ifthen}
\usepackage{bbm}
\usepackage{amsmath}
\newboolean{showcomments}
\setboolean{showcomments}{true} 

\ifthenelse{\boolean{showcomments}}%
{%
    \newcommand{\aasa}[1]{{\color{red} \textbf{Aasa:} #1}}
    \newcommand{\kilian}[1]{{\color{blue} \textbf{Kilian:} #1}}
}%
{%
    \newcommand{\aasa}[1]{}
    \newcommand{\kilian}[1]{}
}

\title{Navigating Uncertainty in Medical Image Segmentation}
\name{Kilian Zepf, Jes Frellsen, Aasa Feragen}
\address{Technical University of Denmark}
\begin{document}
%
\maketitle
\begin{abstract}
We address the selection and evaluation of uncertain segmentation methods in medical imaging and present two case studies: prostate segmentation, illustrating that for minimal annotator variation simple deterministic models can suffice, and lung lesion segmentation, highlighting the limitations of the Generalized Energy Distance (GED) in model selection. Our findings lead to guidelines for accurately choosing and developing uncertain segmentation models, that integrate  aleatoric and epistemic components. These guidelines are designed to aid researchers and practitioners in better developing, selecting, and evaluating uncertain segmentation methods, thereby facilitating enhanced adoption and effective application of segmentation uncertainty in practice.
\end{abstract}
\section{Introduction}

Image segmentation is essential in medical image processing, and leveraging segmentation uncertainty is crucial as we move toward trustworthy and interpretable medical AI. Although progress has been significant, there's an ongoing need for research to specify and understand segmentation uncertainty in new architectures and to include detailed analysis of segmentation tasks and precise categorization of uncertainties as either epistemic or aleatoric~\cite{ditlevsen2008}. 

In this paper, we highlight potential pitfalls in segmentation uncertainty that hinder us as a community to validate or compare models, even with available benchmark data. Focusing on two illustrative cases, we propose a set of guidelines intended to assist future research in designing architectures tailored to their specific problems, facilitating more precise model evaluation and fostering advancement in the field.

\begin{figure}[t]
\centering
\includegraphics[width=1\linewidth]{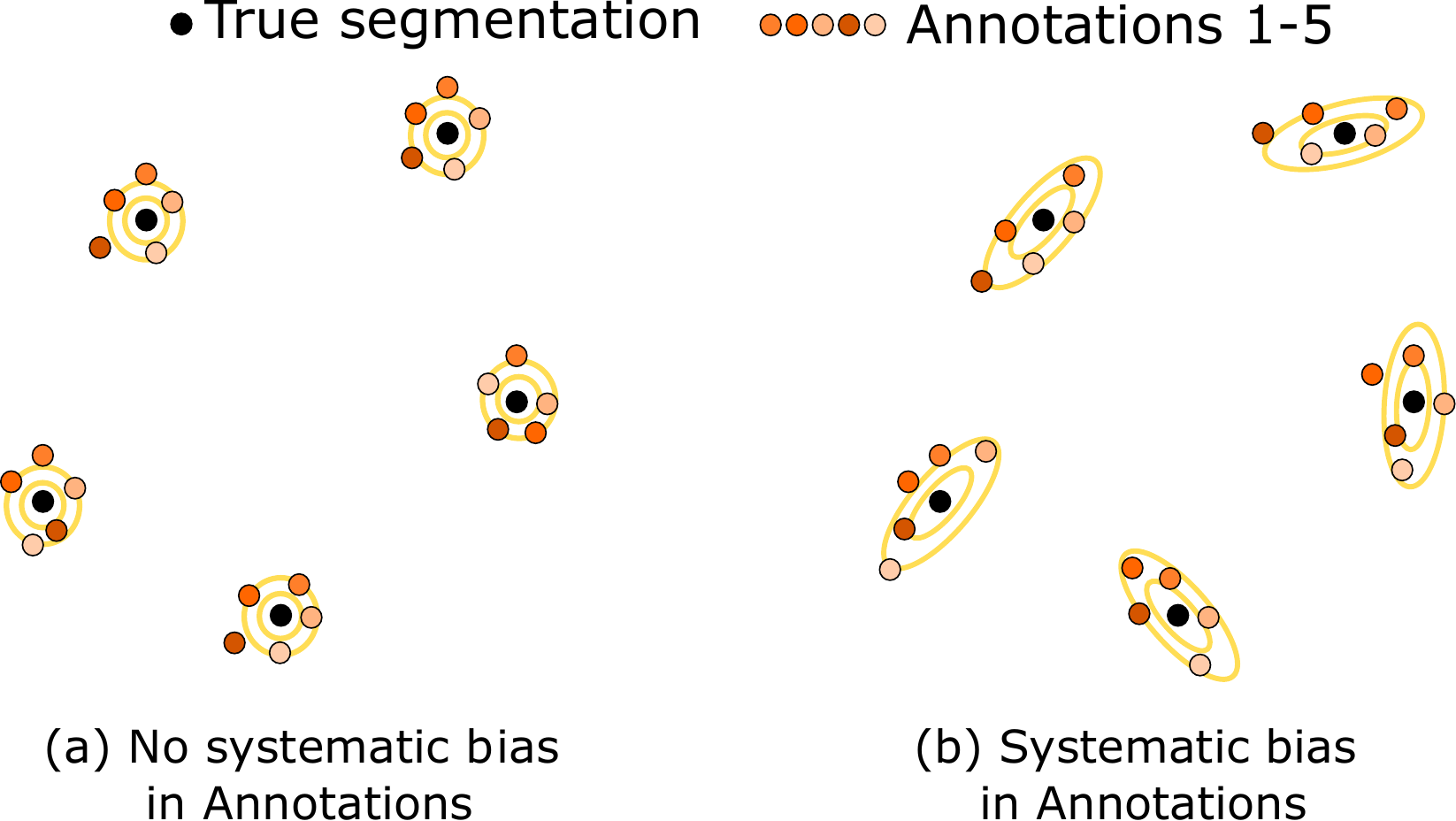}
\caption{Data variation  (aleatoric uncertainty) in segmentation: Illustration of true segmentations $C$ (black dots) together with five expert annotations $A_n^i$ (colored dots). \textbf{(a)} Annotations vary little and unsystematically due to drawing errors $\varepsilon$. \textbf{(b)} Annotations vary systematically, for example caused by ambiguity. The bias $b_n^i$ should influence the method selection.}
\label{fig:modelling}
\end{figure}

Segmentation uncertainty quantification is a catch-all term covering several different estimation problems, which have in common that they seek to attach a measure of uncertainty to predicted segmentations. Consider a dataset where each image has multiple annotated segmentations available, i.e.\@ the dataset consists of pairs $(x_n,A_n^i)$, where $x_n$ is the $n^{th}$ image in the dataset and $A_n^i$ is the $i^{th}$ annotation for this image. Fig.~\ref{fig:modelling} shows a schematic illustration of this situation. In this simple example, we consider the annotated segmentations 
\begin{equation}
A_n^i = f(C, b_n^i, \varepsilon)
\label{eq:eqintro}
\end{equation}
to be a function dependent on the true underlying segmentation $C$, an annotator bias $b_n^i$ and a stochastic annotation error $\varepsilon$ coming from failure to precisely identify and draw the boundary of the object. 

While annotation errors $\varepsilon$ are unpredictable, prior knowledge about annotator biases $b_n^i$ in a dataset might be available such as ambiguity \cite{kohl2018probunet}, differing label styles \cite{zepf2022label} or different skill level of annotators \cite{andreasen2023multi}. This prior information can be used to derive suitable inductive biases for modelling and should be leveraged. However, in cases where $b_n^i$ plays an ancillary role and annotator variation comes down to a drawing error, the need to explicitly model this data variation becomes questionable. 

Recent research has made notable progress in suggesting different architectures for capturing annotator variance, also often named aleatoric uncertainty. A simple but expensive method is an ensemble where each member is trained to reproduce one single annotator. More affordable generative models include multi-head networks~\cite{rupprecht2017learning,lee2016stochastic}, deep latent variable models \cite{kohl2018probunet,monteiro2020,baumgartner2019phiseg,blundell2016ensemble,gal2015bayesian,kohl2019hierarchical} and probabilistic graphical models \cite{NIPS2015markov,kirillov2016joint,markov2018} as well as mean variance networks \cite{monteiro2020}. 

On the other hand, epistemic or model uncertainty expresses what the model does not know and it is high for images the model has not seen during training. This characteristic renders it well-suited for application in out-of-distribution (OoD) detection for downstream tasks. Epistemic uncertainty is intrinsic to the model's parameters and can be reduced as the model learns from more data — when a new image $x$ is added to the training set and the model is updated, the epistemic uncertainty at $x$ declines. Bayesian neural networks, most notably MC-Dropout \cite{gal2016dropout}, and deep ensembles try to quantify this type of uncertainty. Feature space distances and densities provide an alternative to the probabilistic approach in deterministic models by measuring the distance of new samples to training data in feature space; this measure can act as a surrogate for epistemic uncertainty \cite{mukhoti2023deep}.

In the following sections, we explore two practical cases of segmentation uncertainty, illustrated by experiments with aleatoric models such as the probabilistic U-net, Stochastic Segmentation Networks, and annotator ensembles, which are well-established and frequently used methods. Based on our observations, we derive a set of practical guidelines in the Discussion that can be used by future researchers and practitioners in developing uncertain segmentation methods tailored to their needs.

\begin{figure}[t]
\centering
\includegraphics[width=1\linewidth]{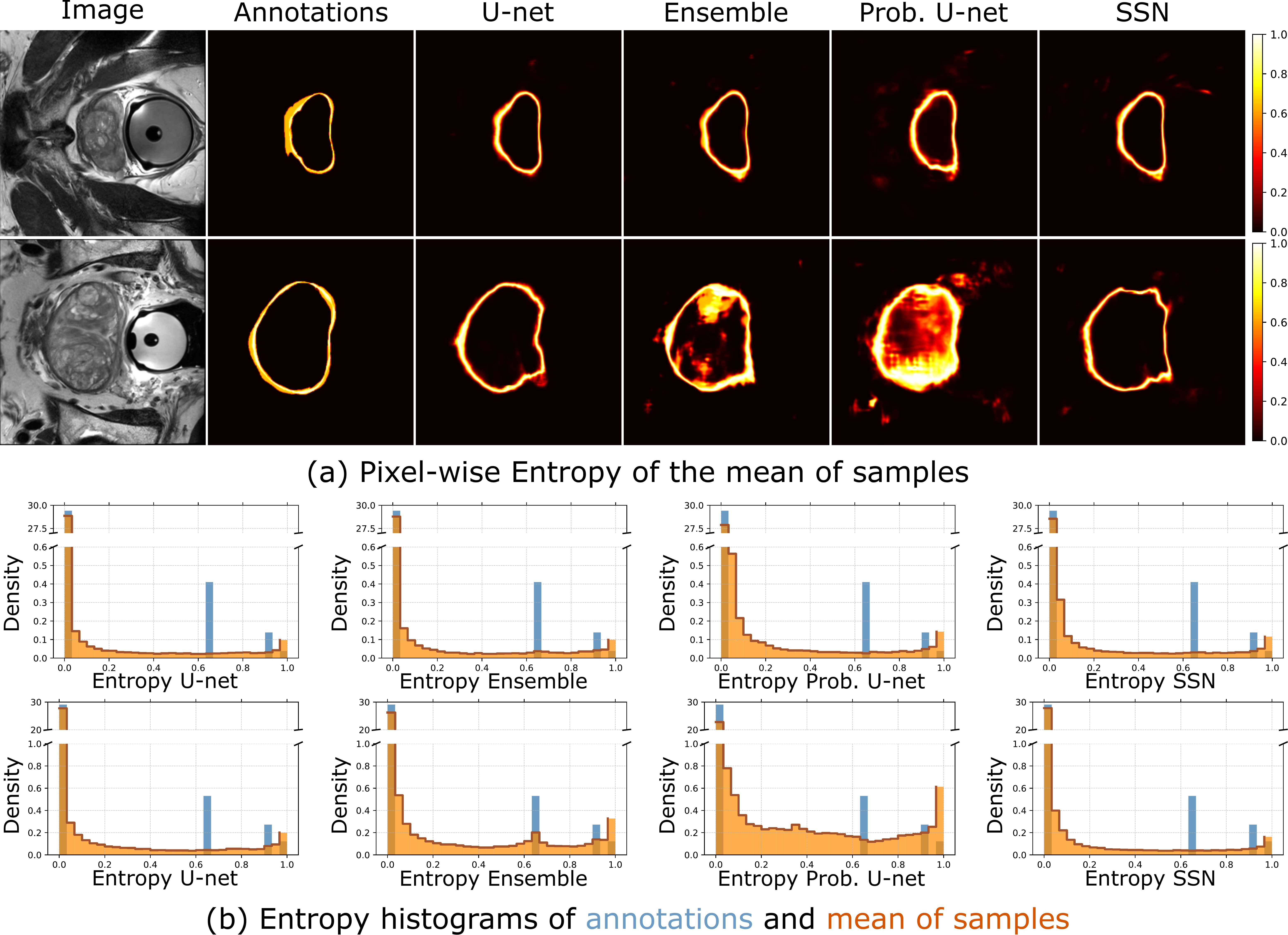}
\caption{A U-net softmax provides alike or better entropy estimates than methods of aleatoric uncertainty. Visual comparison \textbf{(a)} and histogram overlay \textbf{(b)} indicate that the U-net softmax and SSN perform on par while Ensemble and probabilistic U-net might overestimate the variation in the data.}
\label{fig:prostate}
\end{figure}

\section{Prostate segmentation: Estimating annotator variance or noise?}\label{sec:prostate}
We start by considering the first prostate~\cite{becker2019variability} segmentation task from the QUBIQ 2021 challenge\footnote{https://qubiq21.grand-challenge.org}, where each image is segmented by six annotators. As shown by the pixel-wise entropy of the mean over annotations in column 2 of Fig.~\ref{fig:prostate}, this is a case where the manual annotations have very limited variance. We suggest that the main variation in this case comes from the experts’ errors in determining and exactly drawing the organ boundary. This translates to zero annotator bias, and we obtain a model of annotations as 
\begin{equation}
A_n^i = f(C, \varepsilon),
\label{eq:annotation_model}
\end{equation}
where all uncertainty is identification- and drawing error $\varepsilon$. 
In cases where discrepancies are mostly attributed to drawing errors, the focus shifts to the mean segmentation of the annotations $A_n^i$. This mean serves as an effective approximation of the true segmentation. In such a case the variance of the segmentation distribution does not provide additional meaningful insights, as it primarily reflects the inaccuracies in drawing rather than inherent differences in the segmentation process.

An empirical estimate of the mean segmentation is in theory already given by a deterministic model's softmax prediction. In Fig.~\ref{fig:prostate} we compare the entropies for the three most common aleatoric uncertainty methods next to the entropy of a simple U-net softmax. Notably, the ensemble of models, each trained to resemble one annotator, and the probabilistic U-net tend to introduce more variance, even for large sample sizes ($n=1000$). While the entropy profiles of samples from the SSN logit distribution are as close to the ground truth's entropy as the U-net softmax, the SSN comes with more parameters for the additional variance estimation and a costly sampling step when computing the loss. 

We argue that in this case, under the above assumptions, the deterministic model's softmax uncertainty is the most appropriate. The additional complexity in an aleatoric model's architecture and training process does not yield corresponding benefits. In fact, it may pose a potential disadvantage, as it can complicate achieving good model fits. Obtaining an estimator for the true underlying boundary $C$ with as little error as possible might be possible without modelling aleatoric uncertainty beyond a mean softmax prediction.

\section{Validating ambiguity: Detection vs segmentation of lung lesions}\label{sec:lidc}

A special interest has been given to the quantification of aleatoric uncertainty in ambiguous cases of segmentation~\cite{kohl2018probunet,monteiro2020}, where annotators disagree both on the presence of an object in the image, as well as on its boundary. An important dataset for this task is the LIDC-IDRI~\cite{lidcidri} lung lesion dataset, consisting of patches of 2D slices from lung CT, with the associated task of detecting and, when relevant, delineating lesions within these patches. In this dataset, the annotators disagree significantly on whether lesions are present. As a result, the inherent uncertainty pertains both to the detection of lesions, and their boundary delineation. Within the segmentation uncertainty literature, this detection uncertainty has been coined \emph{ambiguity}~\cite{kohl2018probunet}. 

\begin{figure}[t]
  \begin{center}
    \includegraphics[width=0.48\textwidth]{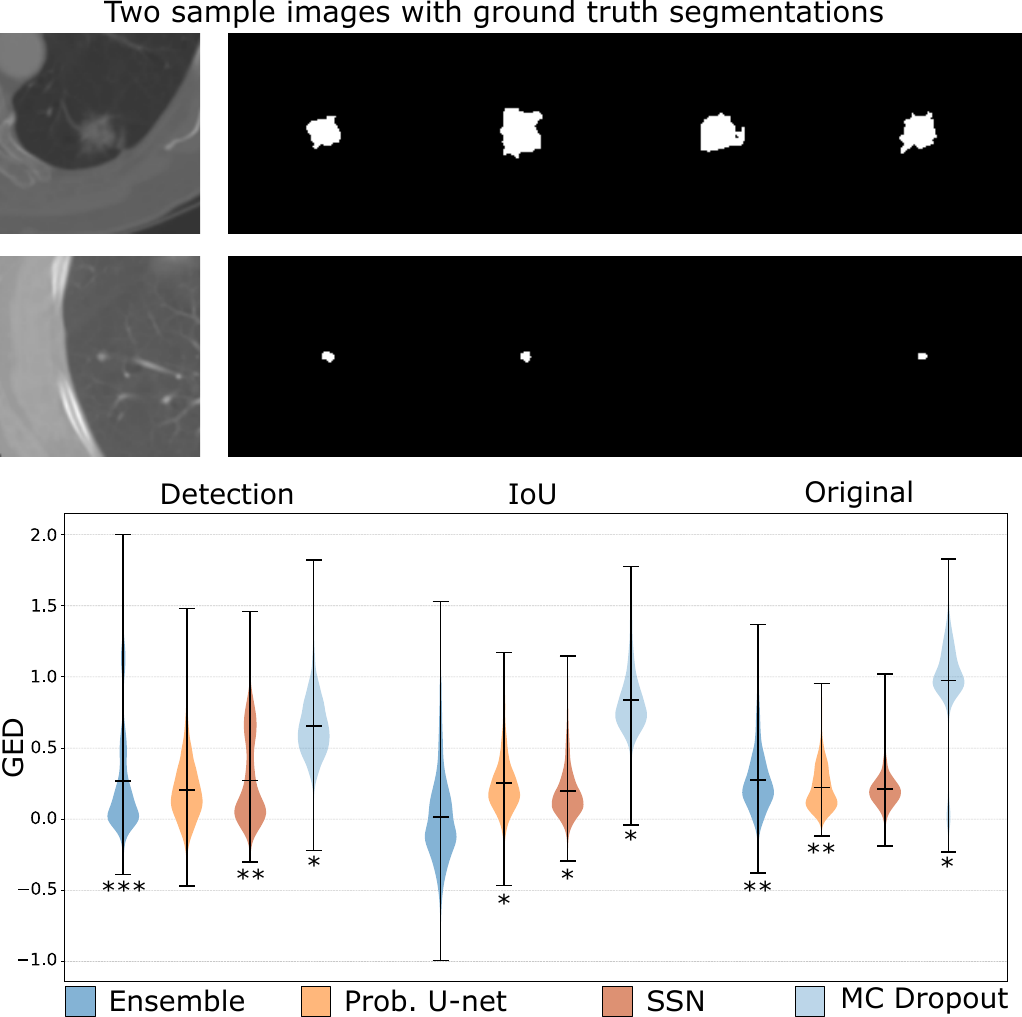}
  \end{center}
\caption{Two samples from the LIDC dataset (top) and violin plots for different GED measures (bottom) for aleatoric uncertainty methods and Dropout. Stars indicate p-values of one-sided Wilcoxon tests between marked and best performing model $*\; p< 2 \times 10^{-308}\; **\; p=5 \times 10^{-63}\; ***\; p=5 \times 10^{-6}\;$. }
\label{fig:lidcresults}
\end{figure}

Aleatoric uncertainty methods are often validated using the generalized energy distance (GED) \cite{baumgartner2019phiseg,monteiro2020,kohl2018probunet} between sampled distributions of ground truth annotations $A \sim P_A$ and sampled model predictions $Y \sim P_Y$ for a given image $x$: 
\begin{align}
\label{eq:ged}
D_{\text{GED}}^2(P_A,P_Y) &= 2\mathbb{E}[d(A,Y)] \nonumber \\
&\quad - \mathbb{E}[d(A,A')] - \mathbb{E}[d(Y,Y')].
\end{align}

The GED comes with the advantage of easy comparability between models by their ability to resemble the annotator distribution. However, as GED scores are usually based on the Intersection over Union (IoU) between pairs of segmentations, that is $d(x,y)=1-\text{IoU}(x,y)$, with $d=0$, if both masks are empty, we show that they are also dominated by the detection, or ambiguity, part of the segmentation task. The reason for this is that when a non-empty segmentation is compared to an empty one, their IoU will always be 0, and this strongly biases the GED towards high values. As a result, the GED scores lack expressiveness for simultaneously evaluating how well uncertainty quantification models estimate detection uncertainty and segmentation uncertainty. We therefore propose to evaluate two additional metrics. Let $A \sim P_{A^*}$ and $Y \sim P_{Y^*}$ the distributions of annotations and model predictions without empty masks respectively, then $D^2_{\text{IoU}}(P_{A^*},P_{Y^*})$ measures the GED only between non-empty masks and predictions. Further we define $D^2_{\text{Det}}(P_{A},P_{Y})$ as in Equation \ref{eq:ged} with a new distance metric $d(x,y) = k(x,y)$ between predictions and ground truths, 
\begin{equation}
  k(x,y) =
    \begin{cases}
      0 & \text{if }(e(x) \land e(y)) \lor (\neg e(x) \land \neg e(y))\\
      1 & \text{if }(e(x) \land \neg e(y)) \lor (\neg e(x) \land e(y)),\\      
    \end{cases}       
\end{equation}
where $e(x)= \mathbbm{1}_{\{x \text{ is empty}\}}(x)$ tests for empty masks.\footnote{$k(x,y)$ in this setting satisfies all metric properties - non-negativity, symmetry, and the triangle inequality — rendering it a valid choice for the use in the Generalized Energy Distance.}

Fig.~\ref{fig:lidcresults} shows violin plots for all three metrics comparing ground truth annotations and three different models, averaged on the test set ($n=1980$). When assessing detection ($D^2_{\text{Det}}$), segmentation ($D^2_{\text{IoU}}$), or detection and segmentation combined ($D^2_{\text{GED}}$), we see different models emerge as the best performing on average. Moreover, the detection seems to dominate the ranking given by $D^2_{\text{GED}}$. We conclude that combined evaluation of detection and segmentation can be suboptimal; additionally using \( D^2_{\text{Det}} \) and \( D^2_{\text{IoU}} \) offers a information that can prove valuable in model evaluation and selection.

\begin{table*}[htbp]
\small
\centering
\caption{Overview of Observations and Guidelines for Segmentation Uncertainty  }
\label{tab:guidelines}
\begin{tabular}{@{}p{0.1\textwidth}p{0.3\textwidth}p{0.5\textwidth}@{}}
\toprule
 & Observation & Guideline \\ \midrule
Inductive Bias & Annotator variation in some datasets with multiple annotators is small and appears as unsystematic drawing errors. & Analyse a priori what type of annotator variation you are dealing with and decide if deterministic methods may suffice. Avoid deterministic models for multi modal annotations (e.g. ambiguity, label styles, ...). \\ \hline
Evaluation & Ambiguity in the data can affect the GED measure; including knowledge of the data into metric design leads to better understanding method performance. & Evaluate the performance of your method in terms of data variation (aleatoric uncertainty) considering your particular data; use metrics that accommodate for potential biases, through for example ambiguity or label styles, to create a clear analysis in what your method excels.\\ \hline
Epistemic gatekeeping and aleatoric inference & Combining aleatoric and epistemic components in segmentation models is necessary, but leads to challenges in interpreting the specific type of uncertainty addressed. & Select methods suited to the uncertainty type you target and try to disentangle them in your method. Evaluate the models' epistemic uncertainty in Out-of-Distribution detection or active learning and aleatoric uncertainty with GED or task-tailored metrics. \\
\bottomrule
\end{tabular}
\end{table*}

\section{Experimental details}

Here we include details on data, implementation and training for the presented experiments. The code is available.\footnote{Release upon publication.}

\textbf{Data.} For the QUBIQ2021 prostate dataset (Sec.~\ref{sec:prostate}) we crop all images to size 640x640 and rescale to 512x512. The training set contained 45 images, with a 7 image test set.
The LIDC-IDRI lung cancer dataset (Sec.~\ref{sec:lidc}) contains 1018 lung CT patches from 1010 patients, with four lesion annotations available per image. We follow the preprocessing of~\cite{kohl2018probunet}, resulting in 128x128 images. Train, validation and test sets contained 8842, 1993 and 1980 images respectively.

\textbf{Implementation and training.} All models are implemented in PyTorch and share a U-net backbone with four encoder/decoder blocks for comparability. Each block contains three convolution layers, bilinear interpolation was used for upsampling. Dropout of $p=0.5$ is used in the lowest level feature map of all architectures, while activated during inference only for the MC dropout U-net. For the probabilistic U-Net we chose a latent space dimension of 6 as in~\cite{kohl2018probunet}. Distribution encoders in the probabilistic U-net are identical with the contraction path of the U-net. All models are trained using the Adam optimizer and binary cross-entropy (BCE) loss. For the probabilistic U-net and the SSN we use the original losses  from \cite{kohl2018probunet} and \cite{monteiro2020}. We used a learning rate of $10^{-4}$ to train U-net, MC dropout and ensemble and $10^{-5}$ for the probabilistic U-net. Further details are available in the code repository. 

\section{Discussion and Conclusion}

We have presented two cases that illustrate the challenges in selecting and evaluating methods for aleatoric uncertainty in segmentation. In the following, we translate our observations in practically applicable guidelines, summarized in Table~\ref{tab:guidelines}, that can be leveraged for both the development of new methods and the selection of a method for a specific task.

Our study of prostate segmentation revealed that annotator variation (aleatoric uncertainty) is often minimal and appears as unsystematic drawing error. Our findings in Section~\ref{sec:prostate} suggest that the methods that are capable of capturing complex annotator distributions have no advantage over a simple U-net softmax prediction. Uni-modal annotator distributions, with low and non-systematic variation, might be well captured by the mean of a deterministic segmentation architecture. However, deterministic models should be avoided when annotations show clear criteria for multi-modality, such as ambiguity or large disagreement. Therefore, the presence of multiple annotations per image can be seen as a necessary, but not sufficient, condition for the use of aleatoric uncertainty methods beyond softmax predictions. We suggest that researchers analyze the type of annotator variation they are dealing with a priori to avoid setting the wrong inductive bias in the first place.

In Section~\ref{sec:lidc}, we demonstrated through the example of lung lesion segmentation that relying solely on the Generalized Energy Distance (GED) measure for model selection can present significant challenges. Three different methods perform best on three different measures, all of which are informative for different subtasks. This finding illustrates that selecting an uncertain segmentation model for a given task should be based on metrics that consider both the particularities of the data (such as ambiguity or label style) and the user's preferences in the trade-off of performance on the involved subtasks. Therefore, in the case of lung lesion segmentation, practitioners might prefer better performance in the detection subtask and accept lower performance in the IoU. However, incomplete evaluation might leave the user unaware that there is an actual trade-off between model performance on different subtasks, and it is therefore crucial to break down strong measures like the GED into more fine-grained and interpretable parts.

In the realm of medical image segmentation, the disentangled combination of epistemic and aleatoric uncertainties in one model is crucial. When a new image is presented to the model, the primary task of the model's epistemic component should be to test whether the image belongs to a familiar (in-distribution) category. This initial epistemic gatekeeping ensures that the model's subsequent aleatoric inference is grounded in its area of expertise (the training data) and alerts the user about possibly inaccurate predictions for unfamiliar data in downstream tasks. 
To keep up with this two-step approach, it is necessary to not only clearly distinguish between methods that quantify epistemic uncertainty (e.g., MC-Dropout or deep Ensembles) and aleatoric variation (e.g., Stochastic Segmentation Networks, Probabilistic U-net, Ensemble of annotators) but also to combine them in one model. While the performance in terms of all three GED measures does not appear competitive, we argue that epistemic components should not be evaluated against aleatoric components. Instead, a joint evaluation of the components should be standard. This requires datasets that exhibit both in-distribution data variation, in the form of multiple ground truth masks per image, and classified and realistic OOD examples.
By defining the roles of epistemic and aleatoric uncertainties clearly, using the former primarily as an OOD detector and the latter for in-distribution predictions, we can cultivate segmentation models that are reliable and clinically impactful. We argue that future research in segmentation should adopt and emphasize this dual-uncertainty framework, building on progress that has recently been made in the formulation of jointly modeling uncertainties \cite{schweighofer2023introducing, mukhoti2023deep}.

With the proposed guidelines for modeling aleatoric and epistemic components, we provide researchers and practitioners with a straightforward and practical framework for effectively employing segmentation uncertainty in medical imaging and other real-world applications.

\subsubsection*{Acknowledgements}  
\small The authors acknowledge the National Cancer Institute and the Foundation for the National Institutes of Health, and their critical role in the creation of the free publicly available LIDC/IDRI Database used in this study as well as the Pioneer Centre for AI, DNRF grant number P1. The work was partly funded by the Novo Nordisk Foundation through the Center for Basic Machine Learning Research in Life Science (NNF20OC0062606). 

\subsubsection*{Compliance with Ethical Standards}
\small This is a theoretical study on open-source datasets for which no ethical approval was required.

\bibliographystyle{IEEEbib}
\bibliography{strings,refs}

\end{document}